\documentclass[lettersize,journal]{IEEEtran}
\usepackage{amsmath,amsfonts}
\usepackage{algorithmic}
\usepackage{algorithm}
\usepackage{array}
\usepackage[caption=false,font=normalsize,labelfont=sf,textfont=sf]{subfig}
\usepackage{textcomp}
\usepackage{stfloats}
\usepackage{url}
\usepackage{multirow}
\usepackage{verbatim}
\usepackage{graphicx}
\usepackage{siunitx}
\usepackage{cite}
\usepackage{epsfig} 
\usepackage{mathptmx} 
\usepackage{times} 
\usepackage{amssymb}  
\usepackage{booktabs}
\usepackage{balance}
\usepackage{hyperref}
\begin{document}

\title{CTBC: Contact-Triggered Blind Climbing for Wheeled Bipedal Robots with Instruction Learning and Reinforcement Learning}

\author{%
  Rankun Li\textsuperscript{*,1,2},
  Hao Wang\textsuperscript{*,2},
  Qi Li\textsuperscript{*,2},
  Zhuo Han\textsuperscript{1,2},
  Yifei Chu\textsuperscript{2},
  Linqi Ye\textsuperscript{\dag,1},
  Wende Xie\textsuperscript{\dag,2},
  and~Wenlong Liao\textsuperscript{2}%
  \thanks{* Indicates Equal Contribution.}%
  \thanks{\dag\ Indicates Corresponding Author.}%
  \thanks{$^{1}$ The School of Future Technology, Shanghai University, 200444 Shanghai, China. {\tt\small rankunli@shu.edu.cn}, {\tt\small yelinqi@shu.edu.cn}}%
  \thanks{$^{2}$ COWAROBOT Co. Ltd., China.}%
}



\maketitle

\begin{abstract}
In recent years, wheeled bipedal robots have garnered significant attention due to their exceptional mobility on flat terrain. However, while stair climbing has been achieved in prior studies, these existing methods often suffer from a severe lack of versatility, making them difficult to adapt to varying hardware specifications or diverse complex terrains. To overcome these limitations, we propose a generalized Contact-Triggered Blind Climbing (CTBC) framework. Upon detecting wheel-obstacle contact, the framework triggers a leg-lifting motion integrated with a strongly-guided feedforward trajectory. This allows the robot to rapidly acquire agile climbing skills, significantly enhancing its capability to traverse unstructured environments. Distinct from previous approaches, CTBC demonstrates superior robustness and adaptability, having been validated across multiple wheeled bipedal platforms with different wheel radii and tire materials. Real-world experiments demonstrate that, relying solely on proprioceptive feedback, the proposed framework enables robots to achieve reliable and continuous climbing over obstacles well beyond their wheel radius. Project page: \url{https://ctbc-for-wheeled-bipedal-robots.github.io/}
\end{abstract}

\begin{IEEEkeywords}
Contact-Based Control, Instruction Learning, Reinforcement Learning, Wheeled Bipedal Robots.
\end{IEEEkeywords}

\section{Introduction}

\IEEEPARstart{T}{raditional} legged robots have demonstrated remarkable agility and adaptability on complex terrain, yet their locomotion efficiency and speed remain comparatively low~\cite{ref1}, making it difficult to satisfy the demands of rapid mobility. Wheeled-legged robots, benefiting from their high energy efficiency over long distances and superior travel speed~\cite{ref2}, have been extensively studied and deployed across various domains. Nevertheless, when confronted with challenging environments such as staircases or uneven surfaces, the inherent limitations of wheeled robots become evident, as they lack the flexibility required to surmount obstacles effectively.

For wheeled-legged robots, tire dimensions exert a decisive influence on the feasibility of stair-climbing. Larger wheels confer a clear advantage; for instance, Simon Chamorro et al.~\cite{ref3} demonstrated that the Ascento robot can ascend a 15 cm stair with wheels of 25 cm radius, but only with deflated tires and limited to climbing a single step.

To overcome the pain points of wheeled-legged robots in complex terrain, we draw inspiration from the contact-triggered reflexes observed in human gait and propose a contact-triggered control framework that synergizes feedforward instruction learning with reinforcement learning. By leveraging privileged information within an asymmetric actor--critic architecture, we develop a biomimetic gait reflex that enables wheeled-biped robots to ascend stairs with ease. In contrast, our method is suitable for wheeled-biped robots with arbitrary wheel diameters and tire types, and has been successfully deployed on robots with 11 cm rubber solid tires and 12.7 cm pneumatic tires. The key contributions of this work are summarized as follows:

\begin{figure}[!t]
  \centering
  \includegraphics[width=\columnwidth]{cover.pdf}
  \caption{We have developed a contact-triggered blind climbing control policy that works for wheeled-legged robots of various tire sizes, enabling them to conquer a variety of challenging terrains.}
  \label{fig:fengmian}
\end{figure}

\begin{enumerate}
\item \textbf{Contact-Triggered RL Task Framework.} We propose a contact-triggered reinforcement learning task framework applicable to wheeled-legged robots of various tire sizes, including the small-tire robot \textit{LimX Dynamics Tron1}, to achieve stair-climbing strategies.
\item \textbf{Feedforward Trajectory Instruction Learning.} By combining instruction learning with reinforcement learning, feedforward trajectories are utilized to teach the robot when to lift its legs appropriately. This approach avoids unnecessary exploration and significantly improves learning efficiency.
\item \textbf{A Universal Framework for Various Wheel Sizes.} This method enables the \textit{LimX Dynamics Tron1} robot to continuously climb 20\,cm stairs (Fig.~\ref{fig:fengmian}), far exceeding its tire radius. It also enables the \textit{Cowarobot R0} heavy-duty wheeled-biped robot to continuously climb 7.5\,cm stairs. The strategy supports both fast movement on flat ground and motion on complex terrain, enhancing the practical adaptability of the robot.
\end{enumerate}

\begin{figure*}[!t]
  \centering
  \includegraphics[width=\textwidth]{framework.pdf} 
  \caption{Overview of our universal contact-triggered blind stair-climbing framework. The overall framework is mainly composed of a state estimator and an asymmetric actor-critic network. For elastic tires, we introduce a contact force sliding window to simulate more realistic contact. When the contact force in the xy direction of the foot exceeds a threshold, it triggers the designed feedforward reference trajectory to guide the robot to lift its leg. After annealing the feedforward trajectory, the method can be zero-shot transferred to the physical robot.}
  \label{fig:side:a}
\end{figure*}

\section{RELATED WORK}

\subsection{RL-Based Legged Locomotion }

In the realm of legged locomotion, model-based optimization techniques such as Model Predictive Control (MPC) and trajectory optimization have long been widely adopted~\cite{ref4}. Nevertheless, the reliance of these methods on accurate and intricate dynamic models often constrains their robustness and generalization. Recently, model-free reinforcement learning has risen to prominence, offering an end-to-end learning paradigm that simultaneously strengthens robustness and markedly enhances generalization across diverse motion-control tasks, positioning itself as a powerful alternative to traditional approaches.

\vspace{-\parskip}
In 2019, Hwangbo et al.~\cite{ref5} introduced reinforcement learning to legged robot control by proposing a policy conditioned on desired velocity that outputs joint-position targets, augmented with an actuator network to enhance motor modeling accuracy. Building directly on this seminal framework, Lee et al.~\cite{ref6} developed a teacher–student architecture that enables robust traversal of hills, steps, and other challenging terrains using only on-board proprioception. Siekmann et al.~\cite{ref7} further demonstrated that an RL policy trained solely on proprioception can drive the bipedal robot Cassie to blindly ascend real-world stairs. Leveraging massively GPU-parallel simulation in Isaac Gym, Rudin et al.~\cite{ref8} trained a locomotion policy with Proximal Policy Optimization (PPO) in just 20 minutes that transfers zero-shot to real hardware. Extending this promising work, Rudin et al.~\cite{ref9} introduced a task formulation based on positional goals: within a strict time limit, the robot must reach a target location, autonomously planning both path and motion to overcome obstacles and complete navigation without additional motion priors.

\subsection{Wheeled-Legged Locomotion on Rough Terrain}

In complex-terrain locomotion, conventional model-based methods often hinge on either simple heuristics that dictate when to walk or when to drive~\cite{ref10}, or on fixed, pre-defined gait sequences~\cite{ref11}. Most policies for legged robots still embed hand-engineered gait patterns~\cite{ref12,ref13} or biologically inspired motion primitives~\cite{ref6,ref14}.

Bjelonic et al.~\cite{ref15} introduced an online gait generator driven by leg availability: when the availability score of any leg drops below a threshold, the leg is automatically switched to swing while the others continue to drive or support; a single MPC parameter set suffices for all gaits, eliminating manual cost-weight tuning. Klemm et al.~\cite{ref16} leveraged non-smooth trajectory optimization to co-solve global motion planning and contact switching for stairs, steps and jumps in one pass, creating a closed perception–control loop and demonstrating continuous stair climbing on the Ascento wheeled-leg platform. Lee et al.~\cite{ref17} trained a quadrupedal wheeled robot with RL to switch on-the-fly between high-speed wheel driving and legged obstacle clearance in response to commands and terrain, enabling robust obstacle traversal. Lee et al.~\cite{ref18} further proposed a fully-integrated end-to-end framework that fuses model-free RL, privileged learning and hierarchical control, allowing seamless transitions between walking and driving for tasks such as table jumping and stair climbing. Chamorro et al.~\cite{ref3} demonstrated that a blind RL policy, operating without vision or localization, allows the Ascento robot to climb 15 cm stairs by relying solely on positional objectives, binary terrain flags, and an asymmetric actor-critic architecture. However, to the best of our knowledge, while the Ascento robot possesses a wheel diameter of 25 cm, both its simulation and hardware experiments only demonstrate the traversal of a single-step stair and lack the capability to ascend stairs with narrow treads.

Currently, a universal obstacle-traversal framework for bipedal wheeled robots with arbitrary wheel sizes remains elusive. In particular, when a robot is required to surmount high steps exceeding its wheel radius without any additional exteroceptive sensing, existing methods often struggle with performance and training convergence.
\section{METHODOLOGY}

As illustrated in Fig.~\ref{fig:side:a}, our universal contact-triggered blind climbing framework is depicted. The following sections systematically detail the training environment, reinforcement-learning task formulation, training pipeline, design of the contact-triggered mechanism, integration of feedforward instruction learning, and the sim-to-real transfer strategy with concrete deployment specifics.

\subsection{Learning Environment} 
\subsubsection{Simulator}
We select Isaac Gym~\cite{ref19} as our training platform because it is specifically designed for reinforcement-learning applications and is equipped with a GPU-accelerated architecture that dramatically increases agent-training speed thanks to its high degree of parallelism. In addition, Isaac Gym supports domain randomization techniques~\cite{ref20}, which improve the robustness of reinforcement-learning agents by introducing environmental variations during training and thus facilitate the transfer of policies to the real world. To validate our sim-to-real pipeline, we also leverage MuJoCo~\cite{ref21} for cross-validation of the trained policies~\cite{ref22}. MuJoCo is renowned for its high-fidelity simulation and is widely used to verify reinforcement-learning policies; a policy that successfully deploys in MuJoCo is generally expected to transfer seamlessly to the physical environment.
\subsubsection{Learning Algorithm}
We adopt PPO with an asymmetric actor–critic architecture~\cite{ref23}.  This variant of the standard actor–critic framework employs separate networks for the actor and the critic, permitting independent updates~\cite{ref24,ref25}.  Building upon this, we further introduce a Multi-Layer Perceptron (MLP) encoder for state estimation~\cite{ref26}.  All training and experiments are conducted on a single NVIDIA GeForce RTX 4090 GPU with 24 GB of VRAM.
\subsubsection{Terrains}
Our environment is structured as an 8 m × 8 m terrain divided into 10 columns: one column of smooth slope, one of rough slope, six columns of stairs, and two columns of discrete obstacles, as shown in Fig.~\ref{fig:terrain}. To progressively increase the curriculum difficulty, the terrain is further split into 10 rows. Generally, the greater the diversity of obstacles encountered during training, the more robust the resulting policy becomes.

\begin{figure}[!t]
  \centering
  \includegraphics[width=\columnwidth]{terrain.pdf}
  \caption{Terrain type. From left to right and top to bottom: smooth slope, rough slope, discrete obstacles, and stairs.}
  \label{fig:terrain}
\end{figure}

\subsection{Task Formulation}
\subsubsection{State}
We adopt an asymmetric actor–critic architecture, so we partition the state into two parts: (i) Observations, which are accessible to both the Actor and the Critic, and (ii) Privileged Information, which is revealed to the Critic only during training.
As summarized in Table~\ref{table:observation}, we explicitly list the observations available to the actor during both training and deployment, as well as the privileged information reserved for the critic at training time.
In particular, the term last actions denotes a composite action vector obtained by a weighted fusion of the raw actions directly output by the network and the actions derived from feedforward trajectory.

\begin{table}[!t]
\centering
\caption{Observation \& Privileged Information}
\label{table:observation}
\setlength{\tabcolsep}{4pt}
\renewcommand{\arraystretch}{1.1}
\begin{tabular}{@{}l l c c c c@{}}
\toprule
Symbol & Description & Units & Coeff. & Size & Noise (\%) \\
\midrule
\multicolumn{6}{@{}l}{\textit{Observation (Actor / Critic)}} \\[2pt]
$\vec{\dot{\theta}}$ & Angular velocity & rad/s & 1.0 & 3 & $\pm20$ \\
$\vec{\gamma}$ & Projected gravity & -- & 1.0 & 3 & $\pm5$ \\
$\vec{q}$ & Joint positions & rad & 1.0 & 6 & $\pm1$ \\
$\vec{\dot{q}}$ & Joint velocities & rad/s & 0.05 & 8 & $\pm50$ \\
$a_{\vec{last}}$ & Last actions & rad \& rad/s & 1.0 & 8 & 0 \\
\addlinespace
\multicolumn{6}{@{}l}{\textit{Privileged Information (Critic only)}} \\[2pt]
$v_x$ & Linear velocity & m/s & 2.0 & 3 & -- \\
$\mu_{\text{contact}}$ & Avg contact forces & N & 1.0 & 6 & -- \\
$h_{\text{height}}$ & Height scan & m & 5.0 & 77 & -- \\
\bottomrule
\end{tabular}
\end{table}

\subsubsection{Actions}
In this study, the robot’s action space has a dimension of 8. These actions correspond to the robot’s individual joints, including both leg and wheel joints. For the leg joints, the action commands are directly used as target positions for low-level proportional–derivative (PD) controllers, i.e., the controllers strive to drive the joints to these preset positions. For the wheel joints, the action vector represents target angular velocities; in other words, the joint motors operate in velocity-control mode, aiming to reach the specified angular velocities.
\subsubsection{Rewards}
Our task-specific rewards are summarized in Table~\ref{tab:rewards_split}. 
The reward function is composed of three main components:

\begin{table}[t]
\centering
\caption{Reward Terms and Classifications}
\label{tab:rewards_split}
\setlength{\tabcolsep}{3pt}
\renewcommand{\arraystretch}{1.15}
\begin{tabular}{@{}l c c@{}}
\toprule
\textbf{Reward} & \textbf{Formula} & \textbf{Coeff.} \\
\midrule
\multicolumn{3}{@{}l@{}}{\textbf{Task Rewards}}\\[2pt]
$\text{Lin. vel tracking x}$ & $\exp\!\bigl(-20\,(v_{\text{cmd},x} - v_{\text{base},x})^2\bigr)$ & 1.2 \\
$\text{Lin. vel tracking y}$ & $\exp\!\bigl(-20\,(v_{\text{cmd},y} - v_{\text{base},y})^2\bigr)$ & 1.0 \\
$\text{Lin. vel tracking x pb}$ & $\frac{\Delta \phi_x}{\Delta t}$ & 1.0 \\
$\text{Lin. vel tracking y pb}$ & $\frac{\Delta \phi_y}{\Delta t}$ & 0.8 \\
$\text{Ang. vel tracking}$ & $\exp\!\bigl(-20\,|\omega_{\text{cmd}} - \omega_{\text{base}}|\bigr)$ & 1.0 \\
$\text{Ang. vel tracking pb}$ & $\frac{\Delta \phi_\omega}{\Delta t}$ & 0.5 \\
$\text{Tracking target pos}$ & $\exp\!\bigl(-2\lVert q-q_{\text{target}}\rVert\bigr)-0.2\lVert q-q_{\text{target}}\rVert$ & 0.8 \\
$\text{Feet air time}$ & $\sum_i \min(t_{\text{air},i},0.5)\,\mathbb{I}_{\text{first contact},i}$ & 2.0 \\
$\text{Feet contact number}$ & $\sum_{i}\bigl[\mathbb{I}_{\text{contact}_i=\text{stance}_i}-1.3\,\mathbb{I}_{\text{contact}_i\neq\text{stance}_i}\bigr]$ & 2.0 \\
$\text{Feet clearance}$ & $\sum_{i}\mathbf{1}_{\text{swing},i}\cdot\mathbf{1}_{h_{\text{min}}<h_i<h_{\text{max}}}$ & 2.0 \\
\midrule
\multicolumn{3}{@{}l@{}}{\textbf{Style Rewards}}\\[2pt]
$\text{Nominal foot position}$ & $\frac{1}{N}\sum_{i}\exp\!\Bigl[-\bigl(\tfrac{(z_i - z_{\text{nom}})^2}{\sigma_z^{2}} + \tfrac{\lVert\mathbf{v}_{\text{cmd}}\rVert^{2}}{\sigma_v^{2}}\bigr)\Bigr]$ & 1.0 \\
$\text{Default pose}$ & $\sum_{j}|q_j-q_{j,\text{default}}|$ & -1.0 \\
$\text{Feet distance}$ & $\max(0,d_{\min}-d)+\max(0,d-d_{\max})$ & -10.0 \\
$\text{Wheel zero velocity}$ & $\exp\!\bigl(-\sum_{j\in\{3,7\}}\mathbf{1}_{\text{swing},j}\,\dot{\theta}_j^{2}\bigr)$ & 0.5 \\
$\text{Same foot x position}$ & $|x_0-x_1|$ & -2.0 \\
$\text{Base height}$ & $|h_{\text{base}} - h_{\text{target}}|$ & -20.0 \\
$\text{Orientation}$ & $\tilde{g}_{x}^{2}+\tilde{g}_{y}^{2}$ & -12.0 \\
$\text{Wheel spin}$ & $\sum_{j}\max(0,\,0.8\,|r\dot{\theta}_j|-\|\mathbf{v}_{\text{foot},j}\|-0.1)$ & -5.0 \\
$\text{Opposite base vel}$ & $\max\!\bigl(0,\,-\text{sgn}(v_{\text{cmd}})\,v_x\bigr)$ & -40.0 \\
$\text{Opposite wheel vel}$ & $\sum_{j\in\{L,R\}}\max\!\bigl(0,\,-\text{sgn}(v_{\text{cmd}})\,\dot{\theta}_j\bigr)$ & -2.0 \\
\midrule
\multicolumn{3}{@{}l@{}}{\textbf{Regularization Rewards}}\\[2pt]
$\text{Lin vel z}$ & $v_{z}^{2}$ & -0.3 \\
$\text{Ang vel xy}$ & $\omega_{x}^{2}+\omega_{y}^{2}$ & -0.01 \\
$\text{Torques}$ & $\sum_{j}\tau_{j}^{2}$ & {\num{-1e-5}} \\
$\text{Dof acc}$ & $\sum_{j}\ddot{q}_{j}^{2}$ & {\num{-2.5e-7}} \\
$\text{Dof vel}$ & $\sum_{j}\dot{q}_{j}^{2}$ & {\num{-1e-5}} \\
$\text{Action rate}$ & $\sum_{j}(a_{j}-a_{j}^{\text{prev}})^{2}$ & -0.01 \\
$\text{Action smooth}$ & $\sum_{j}(a_{j}-2a_{j}^{\text{prev}}+a_{j}^{\text{prev2}})^{2}$ & -0.005 \\
$\text{Collision}$ & $\sum_{i\in\mathcal{I}_{\text{penalised}}}\mathbf{1}_{\lVert\mathbf{F}_i\rVert>1}$ & -50.0 \\
$\text{Feet contact forces}$ & $\max\!\bigl(0,\,\bar{F}_z-F_{\max}\bigr)$ & -5.0 \\
$\text{Dof pos limits}$ & $-\sum_j \max\!\bigl(0,\,|q_j-q_j^{\text{limit}}|\bigr)$ & -2.0 \\
\bottomrule
\end{tabular}
\end{table}

\begin{enumerate}
    \item \textbf{Task rewards}: including velocity-tracking and foot-lifting terms, which ensure the robot moves at the desired speed and follows the prescribed gait pattern.
    \item \textbf{Style rewards}: comprising foot-pose and body-pose terms, which encourage the robot to maintain a natural and stable gait.
    \item \textbf{Regularization rewards}: used to optimize motion smoothness and prevent superfluous joint movements.
\end{enumerate}

Conditioned Reward Formulation: The rewards associated with leg-lifting maneuvers, namely Target Position Tracking, Feet Air Time, Feet Contact Number, and Feet Clearance, are formulated as conditional terms activated solely upon wheel-obstacle contact. This design effectively decouples the locomotion tasks. Specifically, it allows the robot to maintain stable and efficient high-speed wheeled cruising on flat surfaces. Meanwhile, the policy immediately triggers agile leg-lifting behaviors to negotiate obstacles once contact is detected. Consequently, the framework achieves robust obstacle traversal while fully preserving the inherent mobility of the wheeled-bipedal platform.

\subsection{Contact-Triggered Mechanism}
The methodology is primarily inspired by~\cite{ref17}, with the core objective of updating the robot's state and determining the gait phase (stance or swing) based on the horizontal (\emph{xy}-plane) contact forces measured at the feet (Fig.~\ref{fig:Contact-Triggered Mechanism}). We extend this framework with three key enhancements:

\begin{enumerate} \item \textbf{Threshold-based Triggering:} When the contact force on either wheel exceeds a predefined threshold, the feedforward trajectory is immediately activated. This triggers an initial lift of the contacting leg, followed by a synchronized response from the contralateral leg, resulting in a coordinated alternating ascent.

\item \textbf{Sliding-Window Filtering:} 
To address the contact flickering inherent in rigid-body simulators like Isaac Gym, we adopt a three-frame sliding window to aggregate historical contact states, following the approach in \cite{ref3}. To formally define the stable contact condition, we employ an indicator-based sliding window mechanism:
\begin{equation}
    C_{t} = \prod_{i=0}^{2} \mathbb{1}(F_{t-i} > \tau)
\end{equation}
where $C_{t} \in \{0, 1\}$ denotes the binary stable contact state at time $t$. The term $F_{t-i}$ represents the measured contact force magnitude in the $xy$-plane at the $i$-th preceding frame, $\tau$ is the predefined force threshold (e.g., 30N), and $\mathbb{1}(\cdot)$ is the indicator function which equals 1 if the condition is satisfied and 0 otherwise. This filtering mechanism effectively suppresses high-frequency noise while ensuring the reliability of genuine contact triggers.

\begin{figure}[!t]
  \centering
  \includegraphics[width=\columnwidth]{contact-trigger.pdf}
  \caption{Logical architecture of the contact-triggered mechanism.}
  \label{fig:Contact-Triggered Mechanism}
\end{figure}

\item \textbf{Wheel-Leg Synergetic Integration:} The triggering mechanism is tightly coupled with the rolling wheel dynamics, facilitating seamless transitions between continuous rolling and discrete stepping. This integration optimizes the trade-off between energy efficiency and high-performance obstacle traversal. \end{enumerate}

The triggering mechanism determines the lifting sequence by continuously monitoring the contact forces on both feet in real time. For each foot, the system stores the latest three frames of force data and designates the contact as stable contact only if all three frames exceed the threshold.

The prioritized leg-lifting decision logic is structured as follows:

\begin{enumerate} \item \textbf{Asymmetric Contact:} If contact is detected on a single foot only, the system initiates a unilateral lift of that specific leg while the other maintains support or driving (Fig.~\ref{fig:only one contact}).

\item \textbf{Bilateral Contact:} When both wheels encounter obstacles simultaneously (Fig.~\ref{fig:both contact}), the sequence is determined by the reliability and magnitude of the contact signals: \begin{itemize} \item[i)] \textbf{Stability-First:} If only one foot satisfies the stable contact criteria (as defined by the sliding window), that foot is prioritized for lifting to ensure the maneuver is based on a genuine physical obstacle. \item[ii)] \textbf{Force-Dominant:} If both feet exhibit stable contact, the system selects the leg with the greater instantaneous contact force to lift first. This minimizes the resistance torque against the base and facilitates a more agile ascent. \end{itemize} \end{enumerate}

\begin{figure}[!t]
  \centering
  \includegraphics[width=\columnwidth]{only_one_contact.pdf}
  \caption{When either wheel makes contact, only the contacting leg lifts.}
  \label{fig:only one contact}
\end{figure}

\begin{figure}[!t]
  \centering
  \includegraphics[width=\columnwidth]{both_contact.pdf}
  \caption{If both wheels are in contact, the leg with stable contact or the larger contact force is chosen to lift.}
  \label{fig:both contact}
\end{figure}

The mechanism determines the lifting sequence by monitoring bilateral contact forces in real-time. Specifically, the system maintains a buffer of the three most recent force frames for each foot; a "stable contact" is confirmed only if all frames in the buffer consistently exceed the threshold. This logic allows the robot to adapt its locomotion strategy dynamically to ground-truth contact conditions, fostering more natural and stable maneuvers over unstructured terrain.

\subsection{Feedforward Instruction Learning}
The concept of feedforward instruction learning, primarily inspired by~\cite{ref27}, utilizes a baseline gait motion as a heuristic signal to guide policy exploration. In this work, we adapt this mechanism to bipedal wheeled robots by injecting feedforward trajectories exclusively into the hip-pitch and knee-pitch joints. The composite desired joint action $\boldsymbol{a}_t$ is formulated as:
\begin{equation}
    \boldsymbol{a}_t = k_{\text{fb}}\boldsymbol{a}_{\pi}(t) + k_{\text{ff}}(n)\boldsymbol{a}_{\text{ff}}(t),
\end{equation}
\begin{equation}
    a_{\text{ff}}(t) = A \left( 1 - \cos\left( \frac{2\pi}{T}t \right) \right), \quad T=0.6\text{ s},
\end{equation}
where $\boldsymbol{a}_{\pi}(t)$ is the neural-network policy output and $\boldsymbol{a}_{\text{ff}}(t)$ is the reference trajectory. 

The trajectory amplitude $A$ and the 1:2 ratio between the hip and knee joints are chosen to initiate a basic lifting motion. While the lifting clearance is generally proportional to the trajectory amplitude, we purposefully avoid setting an excessively large value. The core objective of the feedforward signal is to serve as a behavioral guide that "seeds" the lifting primitive rather than dictating a precise geometric path. Consequently, once the robot learns the basic intent of lifting its legs, the specific clearance height is autonomously refined and optimized by the RL policy through environmental interaction.

To ensure that the final policy can autonomously execute maneuvers without external guidance during deployment, we implement a linear difference annealing schedule for the feedforward weight $k_{\text{ff}}(n)$:
\begin{equation}
    k_{\text{ff}}(n) = \max \left( 0, k_{\text{0}} - n \cdot \frac{k_{\text{0}}}{N_{\text{ann}}} \right),
\end{equation}
where $n$ is the current training iteration and $N_{\text{ann}}$ denotes the annealing horizon. As training progresses and the policy converges, $k_{\text{ff}}$ gradually decreases to zero. At this stage, the feedforward "scaffolding" is completely removed, and the robot's motion is governed entirely by the trained network policy, ensuring a seamless transition from simulation to real-world deployment.

\begin{table}[b]
\centering
\caption{Domain Randomization}
\label{tab:domain_randomization}
\begin{tabular}{@{}lll@{}}
\toprule
\textbf{Parameter} & \textbf{Range} & \textbf{Unit} \\ \midrule
Payload mass & $[-0.5,\ 2]$ & kg \\
Center of mass shift & $[-3,\ 3] \times [-2,\ 2] \times [-3,\ 3]$ & cm \\
Kp Factor & $[0.8,\ 1.2]$ & N/rad \\
Ka Factor & $[0.8,\ 1.2]$ & N·s/rad \\
Friction & $[0.2,\ 1.6]$ & -- \\
Restitution & $[0.0,\ 1.0]$ & -- \\
Inertia & $[0.8,\ 1.2]$ & -- \\
Motor torque & $[0.8,\ 1.2]$ & N \\
IMU offset & $[-1.2,\ 1.2]$ & -- \\
Default dof pos & $[-0.05,\ 0.05]$ & rad \\
Step delay & $[0,\ 20]$ & ms \\
Push interval & $8$ & s \\
Push vel (xy) & $1.0$ & m/s \\ \bottomrule

\end{tabular}
\end{table}

\subsection{Domain Randomization}
To achieve zero-shot sim-to-real transfer, we introduce a broad set of randomization factors in simulation to model real-world uncertainties and enhance the policy’s generalization ability, as detailed in Table~\ref{tab:domain_randomization}.

In particular, we apply wide-range randomization for \textbf{friction} and \textbf{restitution} to compensate for the simplified contact physics in simulation. A broad friction range enables the blind policy to secure reliable traction across diverse real-world surfaces, while varied restitution coefficients effectively model the tires' passive damping and unpredictable energy dissipation during high-impact collisions with obstacle edges.

\section{EXPERIMENTS}
\subsection{Simulation Experiments}

To quantify the contribution of each component in the proposed CTBC method, we conducted controlled ablation experiments across four distinct configurations. To ensure statistical reliability and eliminate the influence of stochasticity, each method was trained across three independent runs using random seeds ($s=1, 2, 3$). All policies were evaluated after 80,000 iterations under identical simulation environments and hyperparameter setups. The configurations compared are:

\begin{itemize} 
    \item \textbf{CTBC (Proposed)}: Integrates the contact-triggered leg-lifting mechanism with feedforward instructions to achieve coordinated climbing.
    \item \textbf{CTBC w/o Feedforward}: Retains the contact-triggered logic but removes the explicit feedforward trajectory, relying solely on indirect lifting rewards for motion generation. 
    \item \textbf{CTBC w/o Contact-Trigger}: Maintains the feedforward trajectory but lacks the contact-dependent state transition logic, resulting in repetitive or non-adaptive lifting behaviors. 
    \item \textbf{CTBC w/o Both}: Strips away both enhancements to serve as a baseline reinforcement learning policy for performance comparison.
\end{itemize}

For both the CTBC and CTBC w/o feedforward variants, we observed that an initial longitudinal (fore-aft) leg motion serves as a behavioral heuristic that facilitates the learning of the lifting primitive. Consequently, we adopted a two-stage training curriculum: 
\begin{enumerate}
    \item \textbf{Stage I (Exploration):} The policy is trained without constraints on lateral or longitudinal foot placement to encourage the discovery of successful climbing maneuvers.
    \item \textbf{Stage II (Refinement):} Building upon the pre-trained weights from Stage I, we introduce a ``same foot $x$-position'' reward to regularize the fore-aft motion, thereby ensuring a more graceful and stable robot posture.
\end{enumerate}

The comparative performance of the proposed method and its variants is illustrated in Fig.~\ref{fig:terrain level}. As shown in Fig.~\ref{fig:terrain level}, the complete CTBC method consistently outperforms all other variants, reaching the maximum terrain level with significantly higher sample efficiency and stability across different random seeds. 
\begin{itemize} \item \textbf{Impact of Feedforward Trajectory:} Without the feedforward component (CTBC w/o feedforward), the agent lacks effective heuristic guidance during exploration, resulting in a slower progression of terrain levels and a restricted performance ceiling. This underscores that the feedforward signal not only significantly accelerates the policy convergence but is also essential for maximizing the robot's obstacle-clearing capacity. \item \textbf{Impact of Contact-Triggered Mechanism:} The variant lacking the contact-triggered mechanism (CTBC w/o contact-trigger) exhibits poor scalability. Failing to dynamically switch between rolling and stepping modes based on real-time contact states, the robot merely executes repetitive lifting motions. This rigid behavior leads to excessive energy consumption and severely limits its adaptability to varied terrains. \item \textbf{Baseline Performance:} The baseline policy (CTBC w/o both) fails to surmount even the most fundamental obstacles. This outcome demonstrates that neither component alone is sufficient for mastering high-dimensional stair-climbing tasks, further highlighting the necessity of the synergy between feedforward guidance and contact-triggered state transitions. \end{itemize}

\begin{figure}[!t]
  \centering
  \includegraphics[width=\columnwidth]{terrain_level.pdf}
  \caption{Progressive terrain-level training curves for all ablation variants. The solid lines represent the mean performance across three independent random seeds ($s=1, 2, 3$), with the shaded areas indicating the standard deviation. Step height increases from 8~cm to 20~cm while width decreases from 50~cm to 28~cm.}
  \label{fig:terrain level}
\end{figure}

\begin{table}[!t]
\centering
\caption{Success rate (\%) on stairs of increasing height}
\label{tab:stairs_success}
\begin{tabular}{lccccccc}
\toprule
\multirow{2}{*}{Ablation Experiments} & \multicolumn{7}{c}{Step height (cm)} \\
\cmidrule(lr){2-8}
 & 8 & 10 & 12 & 15 & 18 & 20 & 22 \\
\midrule
CTBC (Proposed) & \textbf{100} & \textbf{100} & \textbf{100} & \textbf{98} & \textbf{96} & \textbf{86} & \textbf{70} \\
CTBC w/o feedforward & 96 & 96 & 96 & 92 & 80 & 58 & 38 \\
CTBC w/o contact-trigger & 62 & 60 & 56 & 46 & 18 & 2 & 0 \\
CTBC w/o both & 46 & 34 & 28 & 8 & 4 & 0 & 0 \\
\bottomrule
\end{tabular}
\end{table}

The success rates presented in Table~\ref{tab:stairs_success} further quantify these observations. To ensure the robustness of the evaluation, each success rate was statistically computed across 100 parallel environments with domain randomization fully enabled, accounting for uncertainties in friction, mass, and motor characteristics. While all methods perform reasonably well on 8~cm steps, the advantage of CTBC becomes increasingly pronounced as the difficulty escalates. Notably, our method maintains a high success rate of 86\% at the 20~cm training limit and remains viable (70\%) even at an extreme 22~cm height. In contrast, removing the contact-trigger causes the success rate to plummet to 2\% at 20~cm. This failure occurs because the robot cannot dynamically synchronize its lifting action with actual ground contact, leading to catastrophic instability on higher obstacles.

\subsection{Real-World Experiments}

\begin{figure}[!t]
  \centering
  \includegraphics[width=\linewidth]{10cm_hole.pdf}\\[1mm]
  \includegraphics[width=\linewidth]{16cm_stair.pdf}\\[1mm]
  \includegraphics[width=\linewidth]{20cm_stairs.pdf}
  \caption{Snapshots of real-world experiments on the \textit{LimX Dynamics Tron1} robot. Top: Escaping a 10~cm deep hole through rapid contact-triggered leg lifting. Middle and Bottom: Successfully ascending 16~cm and 20~cm high stairs, respectively, demonstrating robust blind locomotion over varied obstacle heights.}
  \label{fig:10cm hole&16cm/20cm stairs}
\end{figure}

The CTBC policy, operating at a control frequency of 50~Hz, was deployed on the 8-DoF wheeled-legged robot, \textit{LimX Dynamics Tron1}, without any exteroceptive sensing (e.g., LiDAR or cameras). To evaluate the robustness and sim-to-real transferability of the policy, we conducted experiments in two challenging scenarios: hole escape and stair climbing (Fig.~\ref{fig:10cm hole&16cm/20cm stairs}).

\begin{itemize}
    \item \textbf{Hole Escape:} Upon a wheel dropping into a 10~cm deep void, the resultant contact force immediately exceeds the trigger threshold. Despite the discretized 50~Hz control loop, the policy responds by executing a rapid lifting motion, enabling the robot to extricate itself smoothly without losing balance.
    \item \textbf{Stair Climbing:} When encountering 16~cm and 20~cm steps, the policy dynamically determines the lead leg based on the transient contact forces at the impact moment. This facilitates a rapid ascent while maintaining postural stability under blind conditions.
\end{itemize}

\begin{figure}[!t]
  \centering
  \includegraphics[width=\linewidth]{real_pos.pdf}
  \caption{Continuous blind ascent on 20~cm high open-gap stairs. The results demonstrate the policy's precision in handling extreme surface sparsity. The real-time joint data plots confirm the effectiveness of the guidance: the executed trajectories (solid blue) effectively track the prescribed feedforward instructions (dashed red) upon being activated. Notably, the "Contact-Trigger" points mark the precise moments where the agent transitions from rolling to stepping based on impact, enabling stable clearance of consecutive hollow treads.}
  \label{fig:real pos}
\end{figure}

As illustrated in Fig.~\ref{fig:real pos}, the robot successfully ascends a series of 20~cm continuous open-gap stairs. This scenario represents a heightened level of difficulty due to the extreme discontinuity and the hollow structure of the support surfaces. Even in the absence of a solid riser, our method effectively covers these complex terrains: the contact-triggered mechanism provides the necessary precision to initiate lifting only upon impact with the narrow tread, preventing the wheels from falling into the gaps. This robust performance demonstrates that the learned policy can reliably handle repetitive, high-impact locomotion on sparse terrain.

Furthermore, to verify the cross-platform versatility of our framework, we transferred the CTBC policy to the \textit{Cowarobot R0}, a heavy-duty bipedal wheeled-legged robot integrated with a robotic arm and 8-DoF legs. The \textit{Cowarobot R0} presents a stark contrast to \textit{LimX Dynamics Tron1} (which weighs less than 20~kg) with a total mass of 65~kg and significantly different hardware characteristics. Specifically, it utilizes small-diameter (11~cm) solid rubber tires instead of the pneumatic ones used on \textit{LimX Dynamics Tron1}, and its maximum knee joint velocity is constrained to within 2~rad/s. Despite these substantial discrepancies in physical scale, contact dynamics, and actuator bandwidth, the robot successfully achieved continuous ascent of 5~cm and 7.5~cm steps without any fine-tuning (as shown in Fig.~\ref{fig:cowarobot R0}). Such successful deployment on a high-inertia platform, characterized by distinct contact dynamics and stringent kinematic constraints, strongly underscores the exceptional robustness and generalization capabilities of the proposed method.

\begin{figure}[!t]
  \centering
  \includegraphics[width=\linewidth]{5cm_R0.pdf}\\[1mm]
  \includegraphics[width=\linewidth]{7.5cm_R0.pdf}\\[1mm]
  \caption{\textit{Cowarobot R0} ascending 5~cm (top) and 7.5~cm (bottom) steps. Despite the 65~kg mass and low joint velocity ($<2$~rad/s), the robot achieves continuous ascent without fine-tuning, demonstrating the zero-shot transferability of CTBC across platforms with distinct scale and dynamics.}
  \label{fig:cowarobot R0}
\end{figure}

Remarkably, the policy exhibits exceptional generalization: even if the annealing schedule is bypassed and the feedforward signal is abruptly removed, the robot maintains its ability to surmount 20~cm steps. This confirms that the neural network has successfully internalized the lifting maneuver, transitioning from guided exploration to autonomous execution. Furthermore, although the feedforward trajectory was originally designed for a 10~cm lift height, the learned policy successfully scales this behavior adaptively to clear 20~cm obstacles. For even more challenging terrains, the framework remains highly extensible; by increasing the feedforward amplitude and the reward-constrained lifting range, the robot can acquire even more powerful climbing capabilities through re-training.

\section{CONCLUSIONS}

In conclusion, this paper presents a contact-triggered, blind locomotion framework for bipedal wheeled-legged robots, successfully bridging the gap between prescribed feedforward guidance and adaptive reinforcement learning. By internalizing motion primitives through a two-stage training scheme, the robot achieves robust stair-climbing and hole-traversal without any exteroceptive sensing. Experimental validations on both the 20~kg \textit{LimX Dynamics Tron1} and the 65~kg \textit{Cowarobot R0} demonstrate the framework's exceptional cross-hardware versatility. Despite significant discrepancies in physical scale, wheel diameters, and tire materials (pneumatic vs. solid rubber), the policy generalizes effectively to obstacles multiple times the height of the initial guidance.

Nevertheless, certain limitations remain: the extended gait in the first training phase restricts the contact frequency of the rear legs, leading to a persistent kinematic bias where the policy favors leading with the front legs. Furthermore, the current system remains purely blind, limiting its efficiency in complex, unstructured environments. Future work will explore symmetry-breaking rewards to eliminate gait bias and integrate this robust blind policy as a low-level reactive controller within a hierarchical perceptive architecture. This synergy between "blind-reflex" and "vision-based planning" will ultimately empower bipedal wheeled-legged robots with fully autonomous navigation and traversal capabilities in unknown and hazardous terrains.

\balance

\bibliographystyle{IEEEtran}
\bibliography{root}

\newpage

\vspace{11pt}

\vfill

\end{document}